\documentclass[article]{llncs}
\usepackage[T1]{fontenc}

\usepackage{hyperref}
\hypersetup{colorlinks,allcolors=black}
\usepackage[square,numbers]{natbib}
\usepackage{graphicx}
\usepackage{enumitem}
\usepackage{subcaption}
\usepackage{caption}

\begin{document}
\title{Toward a Flexible Metadata Pipeline for Fish Specimen Images\thanks{Supported by NSF-HDR-OAC: Biology-guided Neural Networks for Discovering Phenotypic Traits: 1940233 and 1940322m, NSF HDR-OAC:Imageomics: A New Frontier of Biological Information Powered by Knowledge-Guided Machine Learning: 2118240, and the Institute of Museum and Library Services (IMLS) RE-246450-OLS-20.}}
%
%
\author{Dom Jebbia\inst{1,2,3}\orcidID{0000-0002-9587-8718}\and
Xiaojun Wang \inst{2}\orcidID{0000-0002-2995-9050}\and
Yasin Bakis\inst{2}\orcidID{0000-0001-6144-9440}\and
Henry L. Bart Jr.\inst{2}\orcidID{0000-0002-5662-9444}\and
Jane Greenberg\inst{1}\orcidID{0000-0001-7819-5360}}
\authorrunning{D. Jebbia et al.}
%
\institute{Drexel University Metadata Research Center, Philadelphia, PA 19104 USA\\
\and Tulane University Biodiversity Research Institute, Belle Chasse, LA 70037 USA\\
\and Carnegie Mellon University, Pittsburgh PA 15213, USA\\
\email{djebbia@andrew.cmu.edu};
\email{xwang48@tulane.edu};
\email{ybakis@tulane.edu};\\
\email{hbartjr@tulane.edu}; \email{jg3243@drexel.edu}\\}
\maketitle 
%
%
\begin{abstract}
Flexible metadata pipelines are crucial for supporting the FAIR data principles. Despite this need, researchers seldom report their approaches for identifying metadata standards and protocols that support optimal flexibility. This paper reports on an initiative targeting the development of a flexible metadata pipeline for a collection containing over 300,000 digital fish specimen images, harvested from multiple data repositories and fish collections. The images and their associated metadata are being used for AI-related scientific research involving automated species identification, segmentation and trait extraction. The paper provides contextual background, followed by the presentation of a four-phased approach involving: 1. Assessment of the Problem, 2. Investigation of Solutions, 3. Implementation, and 4. Refinement. The work is part of the NSF Harnessing the Data Revolution, Biology Guided Neural Networks (NSF/HDR-BGNN) project and the HDR Imageomics Institute. An RDF graph prototype pipeline is presented, followed by a discussion of research implications and conclusion summarizing the results.
 \and Digital images \and Biodiversity Collections 
\end{abstract}
\section{Introduction}
Digital technology, cyberinfrastructure, and the full open research movement have enabled new pathways for scientific research. This is particularly true with digital images of scientific specimens. Scientists are able to examine and compare samples on a scale that was not possible in the analog world. Moreover, computational methods enable new modes of inquiry. Although the research opportunities seem endless, researchers face obstacles as they try to sample the correct type of scientific specimen, or develop efficient pipelines to support their work. Many of these challenges stem from metadata quality issues, or simply the absence of metadata, associated with the life-cycle of the digital specimen \cite{elberskirch_digital_2022, gallas_integrated_2010, margaritopoulos_quantifying_2012, schopfel_adding_2013}.\par
A range of metadata challenges in this area became quite apparent as a group of researchers associated with the NSF supported Harnessing the Data Revolution, Biology Guided Neural Networks (HDR-BGNN) project began their work. A key goal of this research is to examine images of fish specimens and their morphological traits via segmentation followed by feature extraction to determine differences among images representing fish from different taxonomic groups. Combining state-of-the-art image segmentation techniques with Phenoscape ontologies for algorithmic analysis \cite{dececchi_toward_2015, manda_using_2015, edmunds_phenoscape_2016, mabee_reasoning_2018, empty_phenoscape_nodate}, researchers could potentially identify undescribed species grouped within currently described species. The collections of images for training neural networks and developing an image-processing workflow revealed many metadata challenges, which led BGNN collaborators at Drexel University’s Metadata Research Center (MRC) and Tulane University’s Biodiversity Research Institute (TUBRI) to develop a flexible, extensible metadata pipeline.\par
This paper reports on the efforts of the MRC-TUBRI collaboration. The next section of the paper provides background context, followed by the underlying goals and objectives. The four-phased approach that framed the work is explained, along with the current RDF-graph prototype model. Finally, the discussion addresses the extensibility of the current model, and the conclusion summarizes the key results.\par
\section{Background context}
Digital technology and data sharing have motivated the development of national and global repositories that provide global access to digital images of biological specimens. Even so, connecting to these repositories and taking advantage of this new infrastructure can be obstructed by a range of challenges associated with metadata and pipeline models \cite{houssos_enhanced_2014, bailey_closing_2021}.\par
\subsection{Open Science Repositories}
Over the past two decades, researchers have supported the proliferation of digital repositories. The growth of these collections has been motivated by a number of key factors, including the open science movement and, most recently, the international embrace of the Findable, Accessible, Interoperable, and Reusable (FAIR) \cite{wilkinson_fair_2016} data principles. 
For the purposes of this paper, it is important to note the role of government policy, which first encouraged and now requires publicly funded data to be made available. These evolving mandates can take several forms.\par
\subsubsection{Europe}
The European Union was the first major government body to develop policy regarding the availability of publicly funded research. It first did so through the Public Sector Information Directive in 2003 \cite{empty_directive_2003}, and later by its 2019 amendment as the Open Data Directive \cite{empty_directive_2019}. The European Commission (EC) has supported these directives by developing infrastructure such as OpenAIRE and Europeana \cite{freire_metadata_2021, freire_aggregation_2019, houssos_enhanced_2014, houssos_implementing_2011, rettberg_openaire_2015, dodero_data_2012, empty_eu-funded_2012}.\par
\subsubsection{United States of America}
 Similarly, in 2013 the U.S. Office of Science and Technology Policy (OSTP) mandated that federal agencies with more than \$100 million in research and development should make their data available within one year of publication \cite{atkins_revolutionizing_2003}. Most recently, in August 2022 the same agency issued a White House supported memo stating that all federally funded research should be available without delay \cite{nelson_desirable_2022}.These policies and similar developments worldwide have created an imperative for academic organizations to make researchers’ data available. They have also encouraged the development of metadata standards that support open data and data interoperability on a global scale.\par
\subsection{Metadata for Open Science and Digital Scientific Specimens}
Open science and open data sharing have motivated the development of many metadata standards, and the adaptation of existing standards. At the general domain level, researchers can apply the Dublin Core (DC) metadata following the extensive list of metadata properties registered at the DCMI Terms namespace \cite{empty_dcmi_2020}. Researchers may also develop a Dublin Core Metadata application profile by integrating metadata properties from other standards with Dublin Core properties. Two well-known examples include the Virtual Open Access Agriculture and Aquaculture Repository VOA3R metadata application profile \cite{diamantopoulos_developing_2011} developed to support the description and reuse of research results in the fields of agriculture and aquaculture as part of a larger federation of open access repositories; and the Dryad metadata application \cite{greenberg_metadata_2009}, which underlies a global repository that publishes research data underlying scientific publications. On a more specific level, there are hundreds of metadata schemes developed for different research domains and types of scientific data. Examples include the Ecological Metadata Language (EML) \cite{michener_creating_2018} for ecology data, the Darwin Core \cite{wieczorek_darwin_2012} (DWC) for scientific museum specimens, and the Data Document Initiative (DDI) \cite{wong_data_2016} for social science research. There are also a wide array of metadata standards associated with the type (e.g., static image, X-ray, moving image), preservation status, and rights specifying data access and usage.\par
The overabundance of metadata standards that can be used to describe scientific data can be both exciting and overwhelming for scientists trying to determine which standards support their data needs. In response to this challenge, various communities have developed directories and registries to help inform decision making and pipeline design. Key examples include the Digital Curation Center’s Disciplinary Metadata Directory \cite{ball_metadata_2016, ball_building_2014}, the Research Data Alliance’s Metadata Standard Directory \cite{ball_building_2014, perez_rdas_2013}, the National Consortium of Biological Ontologies Bioportal  \cite{empty_national_nodate}, and the FAIR Sharing Standards Registry \cite{noauthor_fair_nodate}. These are significant efforts; however, these extensive resources require human examination, which can be daunting. This challenge is quite evident when looking specifically at the metadata for individual specimens. The ‘Life Sciences’ class in the RDA directory includes 32 sub-topic areas. Most of the sub-topics identify five or more metadata standards, and a number of subtopics refer to ten or more applicable metadata standards for any given area. This is also simply within the ‘Life Sciences’ class, and does not include the applicable metadata standards listed in the ‘Physical Sciences \& Mathematics’ and ‘Social \& Behavioral Science’ classes--both of which may include metadata standards that are applicable to physical or other types of scientific specimens. The challenges associated with identifying an appropriate metadata standard further impact metadata pipeline development, data sharing, and the FAIR principles.\par
The FAIR principles motivated this work. FAIR establishes that data should be findable, accessible, interoperable, and reusable. Scientific images, particularly images of specimens housed in digital repositories may be findable and accessible, but the data associated with them is not always interoperable or reusable. These limitations are grounded in metadata  \cite{child_centralized_2022, vlachidis_semantic_2021, leipzig_role_2021, mons_data_2018, courtot_biosamples_2022, batista_machine_2022}. Moreover, they interfere with being able to leverage rich resources for scientific research. One key solution is to develop better metadata pipelines to support FAIR, which is key to the work presented here.\par
\subsection{Metadata Pipelines}
The concept of pipelines denotes a workflow or systems approach to how materials, information, or other types of resources flow from one place to the next, and the stops along the way. Computing and informatics frequently refer to data pipelines to describe the flow of data throughout an information system. A metadata pipeline is, essentially, a type of data pipeline. Metadata pipelines are key to supporting reproducible computational research  \cite{garoufallou_metadata_2022}, and the overall execution of the FAIR principles. A metadata pipeline frequently begins with the harvesting of existing metadata or creation of new metadata in the absence of metadata, followed by the transport of the metadata, often with the associated object, through a series of operations. While a metadata pipeline is intended to support a workflow, the operation is frequently inhibited by inconsistent application of metadata, the absence of key metadata, and conflicting metadata \textemdash all of which impact metadata quality \cite{tsiflidou_tools_2013, park_metadata_2009, park_metadata_2010}. Finally, the identification and implementation of a metadata workflow model presents challenges. Researchers can work with the common workflow language and look at developments, such as the metadata underlying the Open Archival Information System (OAIS) reference model, Digital Asset Management System (DAMS) workflows, or potentially more sophisticated developments, such as the Unified Modeling Language (UML) information model. Another way that may be more comprehensible to researchers is the Resource Description Framework (RDF) model, which underlies the Semantic Web and linked data. All of this has informed the work reported in this paper.\par
\section{Goals and Objectives}
Metadata challenges along with associated metadata model complexities impact the development of successful metadata pipelines. The current circumstance has helped shape the overall goals and objectives that inform our work, the overall goal of which is to develop a flexible and extensible metadata pipeline to support the HDR-BGNN effort. The flexibility allows TUBRI to align the final output of the pipeline with FAIR principles, increasing the impact of the data. Furthermore, the work is also necessary for BGNN to interconnect with the recently established HDR Imageomics Institute. Key objectives shaping our work include:
\begin{enumerate}
  \item Understanding the scope of TUBRI's data flow and metadata needs to accommodate AI research across the BGNN project and the connected Imageomics Institute.
  \item Designing a plan to improve the current metadata pipeline.
  \item Implementing, assessing, and modifying the metadata pipelines as needed.
  \item Demonstrating a proof-of-concept using RDF to align data pipelines and their outputs with FAIR principles.  
\end{enumerate}
Our work is presented in the next section.\par
\section{Designing a Flexible, Extensible Metadata Pipeline}
Our approach to addressing the above objectives and our overall goal was carried out in four phases,identified and discussed here.\par
\subsection{Phase 1: Assessment of the problem}
The process started by evaluating the BGNN metadata lifecycle. First, we determined potential sources of future image collections and what associated metadata elements could potentially be included. Then we considered the future internal needs at TUBRI. Throughout the process, we weighed how these workflows could be restructured to make the dataset useful to the largest audience.\par
This was explored by evaluating the previous data pipelines, workflows, and internal practices at TUBRI and BGNN. Figure 1 demonstrates the fish specimen image pipeline developed by BGNN, as well as the challenges to creating fully automated computational workflows. We contextualized these observations through interviews and collaboration with researchers in the groups. This information was then compared with the practices of other organizations that contribute to the HDR Imageomics Institute \cite{empty_imageomics_2021}, oceanographic data organizations \cite{empty_marine_nodate, empty_introduction_nodate}, and open science repositories.
This assessment identified several deficiencies within the data pipeline. The two most significant were the number of organizations providing collection event metadata and the sparse, irregular conditions of the raw datasets. This created difficulties adapting the ingestion process to normalize metadata with recognized standards (DC, DWC, Exchangeable Image File (EXIF), etc) communicating those choices to users of the dataset. Many organizations have developed various approaches for making their data FAIR, unfortunately those solutions are generally project specific and not often shared in the literature. It was clear that workflows for making datasets FAIR needed to be made more FAIR.\par

\subsection{Phase 2: Investigation of Solutions}
TUBRI researchers determined that there were two approaches to improving the flexibility of the new database structure for BGNN. One is to modify the database schema based on the relational (table-based) database. The other is to switch from the relational database to a document-oriented NoSQL database. Table 1 details the solutions identified during the investigation.\par
The second option is a document-oriented NoSQL database, which is a non-tabular database structure to store the data like a relational database. It offers a fast and flexible schema that enables data models to evolve with frequent changes. The database can use JSON, XML, BSON, and YAML formats to define and manage data.\par
The first approach was chosen because it built upon the relational database structure in use, rather than conceptually redesigning the database structure. Furthermore, this builds upon the semantic interoperability work pursued by earlier collaboration between the MRC and TUBRI on the BGNN project \cite{leipzig_biodiversity_2021, karnani_computational_2022, elhamod_hierarchy-guided_2021, pepper_automatic_2021}. Of the identified techniques, the EAV model was the most adaptable as a database design pattern to restructure the relational databases containing BGNN data. There were also concerns that the JSON and XML data types solutions may cause problems such as poor performance or making the database structure difficult to manage. EAV model was the most abstract of the solutions, but offered methods to redesign the databases to make it more adaptable to new workflows and extensible when ingesting new metadata elements. There are also numerous ways to implement EAV using JSON or XML. EAV implementation can may use an XML column in a table to capture the incomplete information or variable information, while similar principles apply to databases that support JSON-valued columns.\par
RDF was chosen to implement EVA because:
\begin{itemize}
    \item It offers an extensible solution to the ongoing ingestion of new data from disparate sources. 
    \item Major repositories have already adopted some form of RDF, for instance many cultural heritage organizations have adopted it built on the Europeana Data Model, or science data through OpenAIRE.
    \item It makes FAIR principles foundational to the design of data pipelines. 
\end{itemize}
\subsection{Phase 3: Implementation}
Two methods were examined to create an RDF graph to represent the metadata. One is an implementation using Python libraries; the other uses the desktop version of Protégé. Python libraries offer numerous applications and workflows, but Protégé was chosen to create the prototype because the graphical user interface (GUI) provides an interface to directly interact with the graph. Moreover, the Protégé-OWL batch import plug-in offered an efficient, if limited, way to transform spreadsheet data into RDF schema.\par
Team members evaluated the standards in use and chose new control schemes to more accurately describe the metadata. Selected standards are described in Table 3. This included the removal of duplicated, redundant, or deprecated elements. The remaining elements were checked for accurate usage and adjusted accordingly, such as changing {\tt<dwc:AccessConstraints>} to {\tt<dc:accessRights>}. Finally, new schemes were chosen to align the image data with standards used in other photographic applications, for instance, the adoption of standard maintained by Adobe, the International Press Telecommunications Council, and the PLUS Registry, amongst others. The RDF model theoretically allows for the adoption of any standard to normalize data. A rights statement and IRI was also included as an {\tt rdfs:comment} in the graph.\par
Figure 2 represents the RDF graph prototype that was generated. Table 2 lists the previous data containers and the updated database structure and the sources from which the metadata were derived. In the new structure, metadata is grouped into classes based on the kind of metadata in use. For example, {\tt Multimedia} represents the administrative metadata related to the raw image and its capture event. {\tt IQ metadata} refer to the elements generated by BGNN through computational workflows; the training dataset was created by humans and then metadata was generated through segmentation and trait extraction. The {\tt ExtendedImageMetadata} class encompasses image quality metadata for processed images. {\tt Collection event} metadata refers to the specimen data gathered by researchers in the field. {\tt Bach} contains administrative metadata for the final dataset. Each of the top-level updated nodes is assigned an Archival Resource Key (ARK) that serves as a persistent identifier. The ARK associated with {\tt Multimedia} is the parent identifier for the set of images and metadata generated from the workflow.

\subsection{Phase 4. Refinement}
Phase 1 assessed the metadata ecosystem at TUBRI and identified pipeline features that create data bottlenecks and barriers for image processing and segmentation masking. Phase 2 investigated the potential solutions to these identified problems. Phase 3 implemented a prototype RDF model to restructure the BGNN databases. As of this writing, the project is in Phase 4, which synthesizes the results of the previous stages to design more robust and sustainable workflows.\par
Some of the barriers identified in the previous phases include:
\begin{itemize}
    \item Determining which technology or approach is effective and scalable to different collections.
    \item Creating an RDF structure that will enhance the metadata pipeline and make the final datasets more FAIR.  
    \item Designing processes and techniques that are applicable to many different fields, rather than domain specific solutions. 
\end{itemize}
Phase 4 seeks to further investigate and resolve these challenges by:
\begin{itemize}
    \item Creating programmatic workflows that make it easier to create and maintain RDF graphs.
    \item Employing the prototype RDF schema to implement a system that accesses the relational databases as virtual RDF graphs. This allows the query a non-RDF database using SPARQL, access the content of the database as Linked Data over the Web, create custom dumps of the database in RDF formats for loading into an RDF store, and access information in a non-RDF database using the Apache Jena API.
    \item Using a Python wrapper to make the data more accessible to researchers through an application programming interface (API). 
\end{itemize}

\begin{figure}
\begin{subfigure}{.4\textwidth}
\centering
\includegraphics[width=\textwidth]{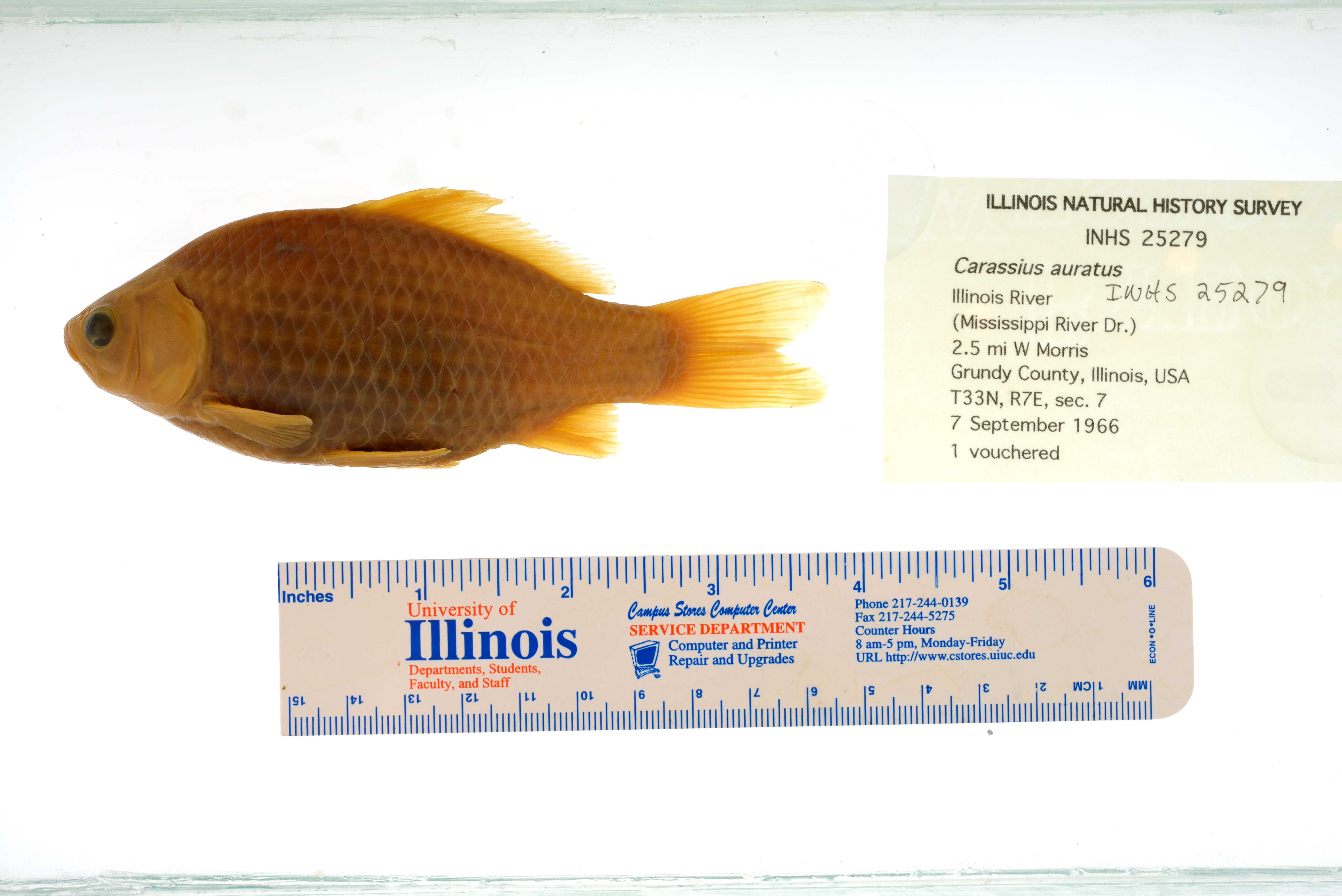}
\caption{A raw image with a ruler and specimen label.} 
\end{subfigure}
\hfill
\begin{subfigure}{.4\textwidth}
\centering
\frame{\includegraphics[width=\textwidth]{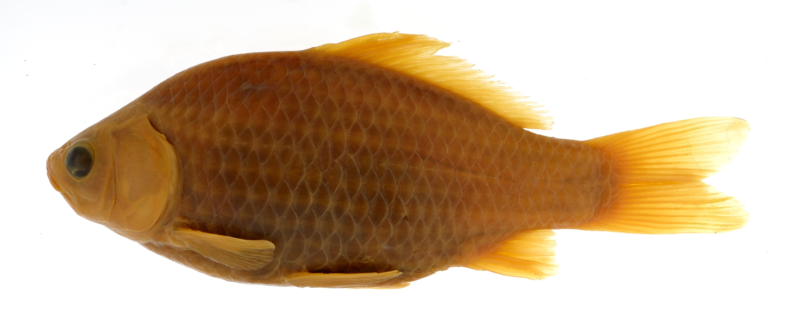}}
\hfill
\caption{A bounding box image is created from the raw image.}
\end{subfigure}
\hfill
\begin{subfigure}{.4\textwidth}
\centering
\includegraphics[width=\textwidth]{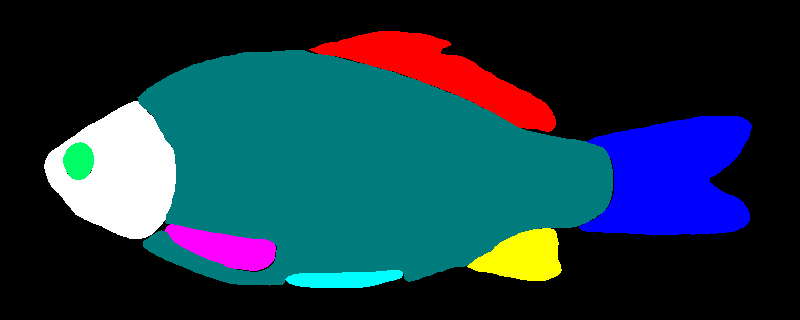}
\caption{A segmentation mask is generated from the bounded image.}
\end{subfigure}
\hfill
\begin{subfigure}{.4\textwidth}
\centering
\includegraphics[width=\textwidth]{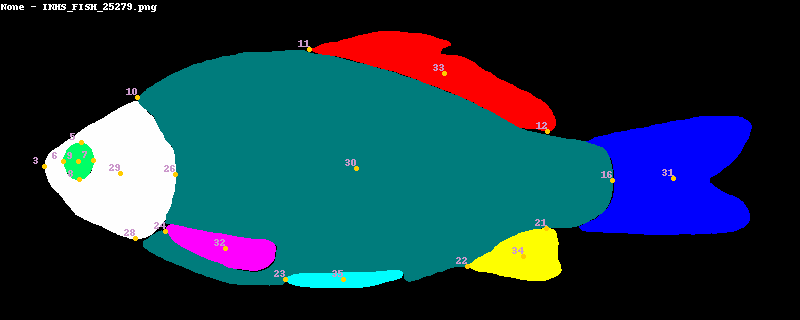}
\caption{Trait features are labeled on the segmentation mask.} 
\end{subfigure}
\caption{The BGNN image processing pipeline featuring a \emph{Carassius auratus} specimen image. Optical character recognition (OCR) is used to extract metadata from the specimen label and validate against the collection event metadata associated with the raw image.}
\end{figure}

\begin{table}
\caption{Relational database solutions.}
\centering
\begin{tabular}{|l|l|}
\hline
Solution & Description\\
\hline
Add columns to tables & Extend existing database.\\
Entity-Attribute-Value (EAV) &  Restructure database using EAV model.\\
JSON data type & Convert database structure to JSON/XML.\\
XML data type & Convert database structure to XML.\\
\hline
\end{tabular}
\end{table}

\begin{table}
\caption{Changes in the database structure.}
\centering
\begin{tabular}{|l|l|l|}
\hline
Original data containers & Updated RDF nodes & Metadata source\\
\hline
Media & Multimedia & Raw image \\
Collection event & Collection event & Specimen\\
ImageQualityMetadata  & IQ metadata & Bounding box image\\
 & ExtendedImageMetadata & Labeled segmentation mask \\
  & Batch & Administrative \\
\hline
\end{tabular}
\end{table}

\begin{figure}
\centering
\includegraphics[scale=0.5]{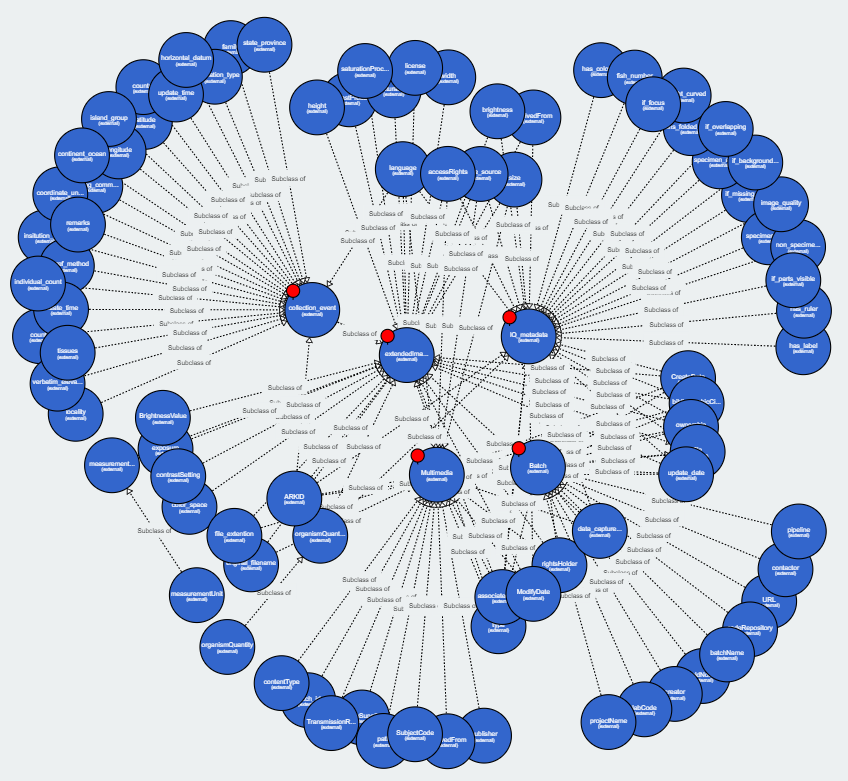}
\caption{A visualization of the RDF prototype created with Protege.} 
\end{figure}
\begin{table}
\caption{Standards added to the RDF prototype.}
\begin{center}
\begin{tabular}{|l|l|l|}
\hline
Standard & Namespace prefix & IRI \\
\hline
Audobon Core & ac & http:\slash \slash rs.tdwg.org\slash ac\slash terms\slash  \\
Camera Raw & crs & http:\slash \slash ns.adobe.com\slash camera-raw-settings\slash 1.0\slash  \\
Darwin Core & dwc& http:\slash \slash rs.tdwg.org\slash dwc\slash terms\slash \\
Darwin Core & dwciri & http:\slash \slash rs.tdwg.org\slash dwc\slash iri\slash  \\
Exchangeable Image File & exif & http:\slash \slash ns.adobe.com\slash exif\slash 1.0\slash \\
IPTC* Core & Iptc4xmpCore & http:\slash \slash iptc.org\slash std\slash Iptc4xmpCore\slash 1.0\slash xmlns\slash  \\
Photoshop & photoshop & http:\slash \slash ns.adobe.com\slash photoshop\slash1.0\slash  \\
Picture Licensing Universal System & plus & http:\slash \slash ns.useplus.org\slash ldf\slash xmp\slash 1.0\slash  \\
Extensible Metadata Platform & xmp & http:\slash \slash ns.adobe.com\slash xap\slash 1.0\slash  \\
Basic Job Ticket & xmpBJ & http:\slash \slash ns.adobe.com\slash xap\slash 1.0\slash bj\slash  \\
XMP Media Management & xmpMM & http:\slash \slash ns.adobe.com\slash xap\slash 1.0\slash mm\slash \\
\hline
\end{tabular}
\end{center}
*International Press Telecommunications Council
\end{table}

\section{Discussion}
\subsection{RDF}
This paper demonstrates how RDF’s flexibility and extensibility can be used to streamline the (meta)data creation process, in addition to providing a database design pattern that can adapt to the changing needs of research investigations. The frameowrk makes research output more FAIR by providing the foundational infrastructure for computational analysis. Specifically, RDF provides a means to express the metadata elements in relation to the resources they describe and to each other, rather than the arbitrary location of the information in a database structure. Investigators spend most of their research time cleaning data \cite{virkus_data_2019}, so pipeline design is an important part of making the final output of a project reusable and ultimately affects the results of later machine learning and neural networks. RDF provides a means of rapidly responding to the changing technical and structural parameters of a project. One major reason for this RDF implementation was to make the database structure able to respond to changing technical requirements, for example, case sensitivity in programming languages. It also provides a means to communicate complex licensing, attribution, and usage rights that accumulate during data reuse.\par
However, the flexibility and extensibility of RDF can present a number of challenges. Although schema can be useful in database design \cite{kalogeros_document-based_2020}, if RDF is implemented without consideration of FAIR principles the resulting database structure may inhibit its ability to link data through the Semantic Web \cite{mons_cloudy_2017}. One way that this could occur is by applying an RDF structure to a database without curating the standards defining the data. Because data collection is laborious and time-consuming, as research approaches evolve it can be difficult to maintain structured data. RDF can make this data findable and accessible to some degree, however without a contextual data model scientists may need to further analyze and clean the data, contact the original investigators for further information, or guess as to the details of the original investigation. This can severely impact quality and reproducibility. Furthermore, the last few decades of digitization efforts by galleries, libraries, archives, and museums have produced corpora of semi-structured historical data relating to every domain of science documented before the digital age. These datasets are important to climate and biological scientists as they attempt to understand climate change and biodiversity \cite{soltis_digitization_2017, nordling_scientists_2019, fordham_using_2020, rockembach_climate_2021}. However, the people who created these early datasets had no idea how the data they were collecting could be used by others in the future. The cost of data management and geopolitical remnants of colonialism are also significant barriers to making data FAIR \cite{brunet_historical_2014, nordling_scientists_2019, sterner_fair_2022}. RDF can help make these datasets findable and accessible if the circumstances allow, however significant curation is necessary to make them interoperable and reusable by machines.\par
\subsection{FAIR Pipelines for Open Science Repositories}
One of the desired outputs of the BGNN project is a dataset of processed images, segmented masks, and rich metadata for others to reuse in future studies. The RDF database structure makes it easy to manage and update data structures, resulting in the ability to accept new metadata elements and adjust them as the requirements of the project evolve. The workflows that RDF makes possible improve the quality and quantity of metadata associated with the images in the dataset. This helps align the research output with FAIR, while also designing pipelines that can be adopted by other studies interested in building FAIR aligned workflows. Although researchers conceptually understand metadata, it is difficult to stay up-to-date with the technical and practical nuances of metadata creation, even among metadata professionals \cite{chuttur_perceived_2013}. As technology becomes more sophisticated and metadata standards proliferate, there is a growing need for researchers to use adaptable schemes in their pipelines to make their data interoperable with machines. Open science mandates from governments and funders will further encourage scholars to house research datasets in open repositories. Data repositories have a role encouraging the adoption of RDF schemes that will make curated data more FAIR.\par 
\subsection{Future Research}
As discussed in section 4.4, the Phase 4 Refinement will continue to refine the RDF protype. The prototype has already been used to construct a demo REST API to interact with the BGNN dataset \cite{tubri-github_tubri-githubbgnn_api_2022}. The API provides both a GUI to search the dataset by genus or ARK identifier, as well as command line access using {\tt cURL} and {/tt Wget}. An API call will download a zip file that contains:
\begin{itemize}
    \item CSV files containing the metadata associated with each image.
    \item XML files containing he metadata associated with each image.
    \item A text document with the preferred citations.
    \item An OWL file containing the RDF graph.
\end{itemize}
The future focus of the the MRC-TUBRI collaboration is to continue refining the RDF model and testing the the API. Further investigation into different modes of RDF adoption for data management and metadata creation is needed to understand other database implementations using RDF automatic workflows for creating and managing knowledge graphs. \par 

\section{Conclusion}
This paper reports on an initiative targeting the development of a flexible metadata pipeline through a collaborative effort involving the MRC-TUBRI. A key contribution is a four-phased approach covering the 1. Assessment of the Problem, 2. Investigation of Solutions, 3. Implementation, and 4. Refinement. The other key contribution is the presentation of the RDF graph prototype. The work presented has been applied to over 300,000 digital images of scientific specimens, specifically fish images, drawn from multiple collections. While we are in the early stage of the RDF graph prototype, the biologist and computer scientists are finding that the workflow and the model expedites their work to service the larger BGNN team in seeking image samples for training the bio-generated neural network. Our next steps include extending our model to other images in the Imageomics institute, given the broad applicability of this work.\par
As already stated, open data sharing has motivated development of many metadata standards, and a range of metadata models. Indeed these standards aim to ensure smooth operations, whether the goal is resource discovery, support for other aspects of FAIR, or integrating into an AI operation. Metadata is a form of data intelligence, and significant time and money are involved in developing, reviewing, endorsing and implementing standards. With respect to the work reported on in this paper, the initial metadata spreadsheet reviewed was loosely structured around the database containers where the various elements were stored as a result of the pipeline structure. This metadata was roughly organized by metadata creation or modification date. Our four-phased approach and adoption of RDF presents a proof of concept for expressing the metadata elements and their relationship to each other rather than the specific location of the data. This work has helped the team achieve a flexible and extensible metadata pipeline. Our overall conclusion is that the RDF graph prototype and our 4-phased approach is flexible and extensible to the wider variety of analysis of a full range of images being examined in the Imageomics institute. In addition, the proof-of-concept is applicable to other metadata pipelines, and supports computational analysis.\par
\subsubsection{Acknowledgments} We thank the Integrated Digitized Biocollections (iDigBio), Global Biodiversity Information Facility (GBIF) and MorphBank data repositories, and the curators of the fish collections in the Great Lakes Invasives Network -- Field Museum of Natural History, Illinois Natural History Survey, J. F. Bell Museum of Natural History, Ohio State University Museum of Biological Diversity, University of Michigan Museum of Zoology, and University of Wisconsin-Madison Zoological Museum – for sharing images of their fish specimens with us.  We also thank Anuj Karpatne and team at Virginia Tech University who developed and trained the fish feature segmentation ANN component of the workflow, Joel Pepper for automated image quality feature extraction workflow and Bahadir Altintas for developing automated landmark extraction workflow.

\

\bibliographystyle{splncs04}
\bibliography{biblio}
\end{document}